\documentclass[10pt,twocolumn,letterpaper]{article}

\usepackage[accsupp]{axessibility}
\usepackage{cvpr}              
\usepackage{listings}
\usepackage{microtype}

\usepackage[dvipsnames]{xcolor}

\definecolor{cvprblue}{rgb}{0.21,0.49,0.74}
\usepackage[pagebackref,breaklinks,colorlinks,citecolor=cvprblue]{hyperref}
\usepackage{bbding}
\usepackage{multirow}

\def\paperID{5829} 
\def\confName{CVPR}
\def\confYear{2024}

\title{Density-guided Translator Boosts Synthetic-to-Real Unsupervised Domain Adaptive Segmentation of 3D Point Clouds}

\author{Zhimin Yuan$^{1}$\hspace{1mm}
Wankang Zeng$^{1}$\hspace{1mm}
Yanfei Su$^{1}$\hspace{1mm}
Weiquan Liu$^{1}$\hspace{1mm}
Ming Cheng$^{1}$\footnotemark[1]\hspace{2mm}
Yulan Guo$^{2}$\hspace{1mm}
Cheng Wang$^{1}$\\
$^1$  Fujian Key Laboratory of Sensing and Computing for Smart Cities, 
Xiamen University\\
$^2$ National University of Defense Technology\\
}

\begin{document}
\maketitle
\footnotetext[1]{Corresponding author.}
\begin{abstract}

3D synthetic-to-real unsupervised domain adaptive segmentation is crucial to annotating new domains. Self-training is a competitive approach for this task, but its performance is limited by different sensor sampling patterns (i.e., variations in point density) and incomplete training strategies. In this work, we propose a density-guided translator (DGT), which translates point density between domains, and integrates it into a two-stage self-training pipeline named DGT-ST. First, in contrast to existing works that simultaneously conduct data generation and feature/output alignment within unstable adversarial training, we employ the non-learnable DGT to bridge the domain gap at the input level. Second, to provide a well-initialized model for self-training, we propose a category-level adversarial network in stage one that utilizes the prototype to prevent negative transfer. Finally, by leveraging the designs above, a domain-mixed self-training method with source-aware consistency loss is proposed in stage two to narrow the domain gap further. Experiments on two synthetic-to-real segmentation tasks (SynLiDAR $\rightarrow$ semanticKITTI and SynLiDAR $\rightarrow$ semanticPOSS) demonstrate that DGT-ST outperforms state-of-the-art methods, achieving 9.4$\%$ and 4.3$\%$ mIoU improvements, respectively. Code is available at \url{https://github.com/yuan-zm/DGT-ST}.

\end{abstract}    
\section{Introduction}
\label{sec:intro}


\begin{figure}[t]
  \centering
  \includegraphics[width=1\columnwidth]{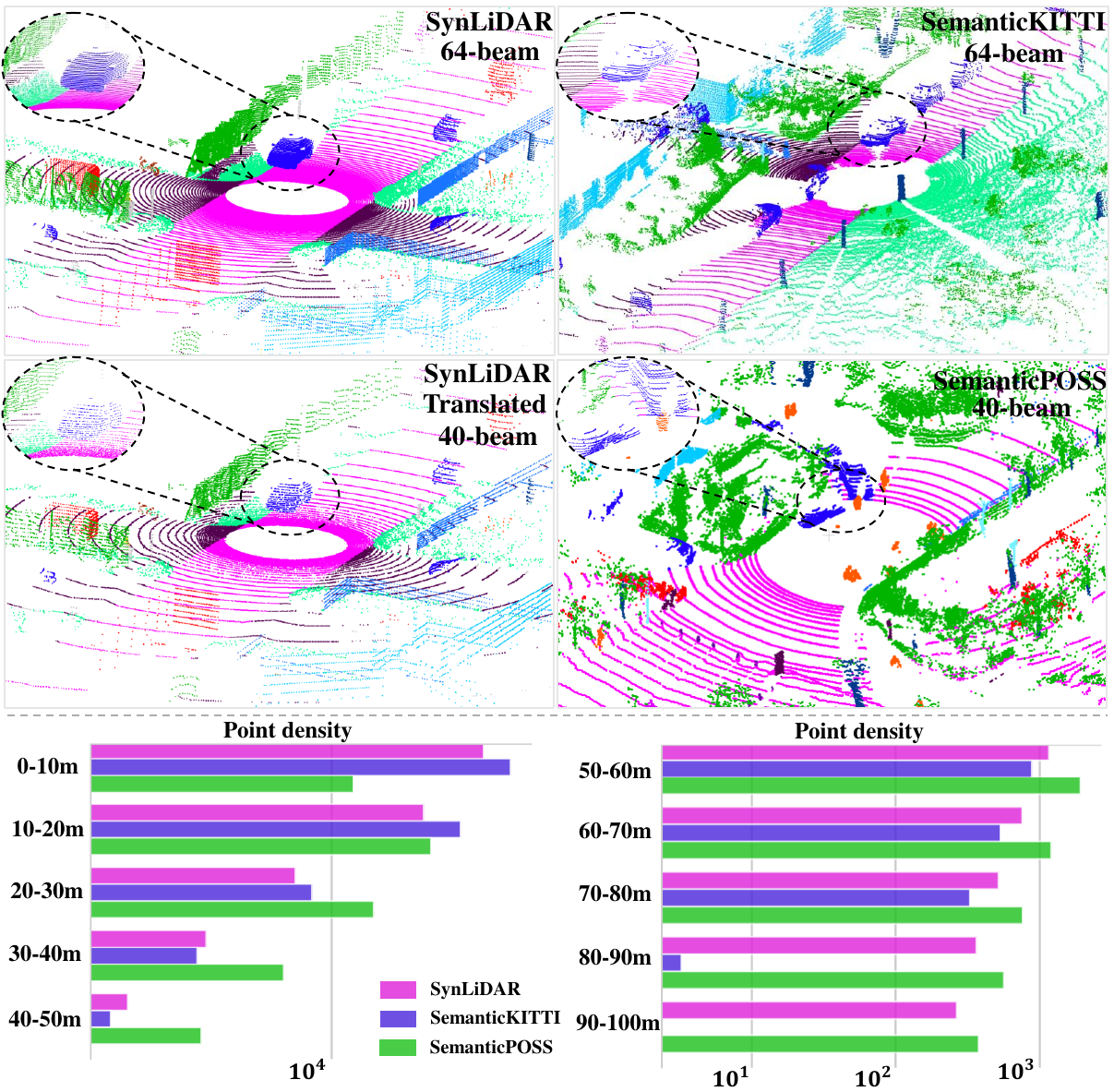}
   \caption{Distinct sampling patterns between synthetic and real-world scans. The synthetic scan (upper left) is integral and clean, whereas the real-world data (upper and middle right) contains unexpected and irregular noise. DGT enhances the realism of synthetic scan (middle left). Point densities of three datasets at various distances from the LiDAR center are shown at the bottom.
   }
   \label{fig:1}
\vspace{-0.5cm} 
\end{figure}
3D point cloud segmentation is crucial owing to its diverse applications, \eg, autonomous driving and robotics. Although supervised methods~\cite{hu2020randla,ando2023rangevit,lai2022stratified,zhu2021cylindrical,yan20222dpass} have made substantial strides, they need costly human-annotated data. In contrast, we can obtain massive synthetic labeled data from simulation platforms. However, the domain gap between the synthetic (source) and real-world (target) data makes
training directly using synthetic data infeasible.
One alternative approach is unsupervised domain adaptation (UDA), which transfers the learned source domain knowledge to make models perform better on the unlabeled target domain.

UDA seeks to acquire invariant knowledge across domains. As shown in \cref{fig:1}, the sampling pattern mismatch between different sensors is the primary cause of the 3D domain gap, directly resulting in distinct point densities (\ie, the number of beams and the point number per beam) between domains. Besides, the synthetic data is integral and clean, whereas real-world data contains varying degrees of noise. Contemporary 3D synthetic-to-real UDA segmentation methods can generally be categorized into two groups: (1) Adversarial training~\cite{yi2021complete,jiang2020lidarnet,zhao2021epointda,yuan2022category,yuan2023prototype,li2023adversarially}, which adopts a discriminator to ensure features/predictions of the segmentor domain-invariant. The limitation of this line of approaches lies in the tendency of aligning the distributions of the two domains as a whole (\ie, global-level alignment), resulting in suboptimal performance; (2) Self-training~\cite{kong2023conda, saltori2022cosmix, xiao2022polarmix,shaban2023lidar}, which usually performs better than the former. It adopts the data mixing techniques to construct an intermediate domain and uses the high confidence pseudo-labels to gradually learn the target domain knowledge. However, it heavily relies on a well-initialized model to provide confident pseudo-labels.

After dissecting the existing works, we find two key problems not adequately addressed. (1) \textit{Input data}: although the existing approaches~\cite{xiao2022transfer,li2023adversarially} notice the problem of different sampling patterns, they strive to generate target-like source data and fulfill feature alignment simultaneously in the unstable adversarial training,
which are difficult to reconcile. (2) \textit{Pretrained model}: the existing 3D UDA self-training methods~\cite{xiao2022polarmix,saltori2022cosmix,kong2023conda} completely overlook the importance of the well-initialized pretrained model. They directly employ the model trained on the source domain as the pretrained model, and the unsatisfactory pseudo-labels significantly limit their performance. Drawing inspiration from the 2D UDA counterparts~\cite{zhang2021prototypical,zhang2019category}, we conjecture that adversarial training warm-up is essential for 3D self-training.



In this work, we propose a LiDAR scan translation strategy and integrate it into a two-stage self-training pipeline named DGT-ST to address the two issues above respectively. For (1), to solve the distinct point density across domains, we propose a non-learnable \textit{density-guided translator} (DGT). It is a statistical-based module that narrows the domain gap at the input level by generating the other domain-like scans for each domain. Specifically, we divide a scan into discrete areas and
use point density of each area to determine the location and number of points to be discarded, matching the density of the corresponding area in the other domain. 



For (2), to provide a well-initialized model for self-training, we propose a \textit{prototype-guided category-level adversarial network} (PCAN) in DGT-ST stage one. We use prototypes to dynamically measure
the aligning confidence of points, and propose a self-adaptive reweighting strategy to reduce the impact of adversarial loss on those well-aligned points. This strategy effectively prevents negative transfer. We leverage this well-initialized model to perform self-training in stage two. During the self-training process, we propose a \textit{source-aware consistency LaserMix} (SAC-LM) to learn the source knowledge from the target data,
which enforces the segmentor to give consistent predictions on the target scans and the corresponding scans translated by DGT. Moreover, we employ the teacher-student training strategy to provide robust pseudo-labels and extend LaserMix~\cite{kong2023lasermix} into UDA segmentation to fully utilize the spatial prior of both domains and bridge the domain gap.

Our main contributions can be summarized as follows:
\begin{itemize}

\item We propose a statistical-based density-guided translator (DGT) that directly bridges the domain gap at the input level. Based on DGT, we propose DGT-ST training pipeline for 3D synthetic-to-real UDA segmentation.

\item 
We design PCAN and SAC-LM to constitute the two stages of DGT-ST, respectively. For the former, we use the prototype to perform category-level adversarial alignment to prevent negative transfer and provide a well-initialized model for subsequent self-training. For the latter, we propose a self-training method that extracts source knowledge from target data, further improving the domain-invariant feature extraction power of the segmentor. 


\item Extensive experiments on two synthetic-to-real tasks verify the effectiveness of DGT-ST, which outperforms the state-of-the-art UDA approaches by a large margin.

\end{itemize}

\section{Related work}
\label{sec:rela_w}
\textbf{Point clouds semantic segmentation.} 3D semantic segmentation aims to give each point a semantic label. PointNet~\cite{qi2017pointnet} is the pioneering work in point-based~\cite{qi2017pointnet++,hu2020randla,yan2020pointasnl,qian2022pointnext,zhao2021point,lai2022stratified,park2023self} methods, employing multilayer perceptrons to extract point features. Then, numerous point-based methods have been proposed and achieved outstanding results. However, these methods generally demand extensive computational resources, which makes them difficult to deploy in practical applications. In contrast, some approaches~\cite{wu2018squeezeseg,wu2019squeezesegv2,cortinhal2020salsanext,ando2023rangevit,kong2023rethinking} project the 3D point clouds into 2D grids and leverage the 2D network to perform segmentation tasks. These approaches are efficient as they eliminate the demand for sampling and neighbor search operations. However, the loss of 3D geometric/topological information limits their performance. Currently, the prevailing approach is the voxel-based method~\cite{xu2021rpvnet,zhu2021cylindrical,lai2023spherical,tang2020searching,choy20194d}. These methods convert points into voxels and use sparse convolutions to extract geometric relationships between voxels. Due to its efficiency and promising result, we select MinkUnet~\cite{choy20194d} in this work.

\textbf{Point clouds UDA semantic segmentation.} UDA segmentation aims to use the labeled source data and unlabeled target data to train a model to perform well on the target. Following the spirit of the 2D counterparts, the dominant 3D UDA segmentation methods can be roughly divided into two groups: adversarial training and self-training. ePointDA~\cite{zhao2021epointda} employs CycleGAN~\cite{zhu2017unpaired} to render dropout noise for domain alignment explicitly. Complete $\&$ Label~\cite{yi2021complete} attempts to complete the two different domain data to a canonical domain, which bridges the domain gap at the input level. LiDARNet~\cite{jiang2020lidarnet} simultaneously extracts domain-shared and domain-private features while employing two discriminators that jointly adapt for semantic and boundary predictions. PCT~\cite{xiao2022transfer} employs two generators and discriminators to translate the point cloud appearance and sparsity, respectively. ASM\cite{li2023adversarially} designs a novel learnable masking module to mimic the pattern of irregular noise and mitigate the domain gap. However, the global-level adversarial alignment can easily cause negative transfer. Two category-level adversarial alignment methods~\cite{yuan2022category,yuan2023prototype} are proposed and show encouraging performance to solve this problem.


Self-training~\cite{kong2023conda, saltori2022cosmix, xiao2022polarmix,shaban2023lidar} is another line for this task, which leverages pseudo-labels to learn target knowledge gradually. ConDA~\cite{kong2023conda} proposes an image concatenation-based framework for interchanging signals from both domains. PolarMix~\cite{xiao2022polarmix} proposes cutting, editing, and blending of two domain scans to enrich the data distribution for alignment. CosMix~\cite{saltori2022cosmix} proposes a domain-mixing strategy that harnesses semantic and structural information to reduce the domain gap. They adopt a mean-teacher~\cite{tarvainen2017mean} paradigm to get robust pseudo-labels. However, these methods employ a model trained only on the labeled source domain as the pretrained model, and the unreliable pseudo-labels significantly limit their performance. LiDAR-UDA~\cite{shaban2023lidar} proposes a two-stage method that exploits random discarding source domain beams to obtain a pretrained model, and utilizes the temporal consistency of consecutive frames to generate reliable pseudo-labels. However,
it does not take into account the number of points per beam.






\section{Methodology}
\label{sec:method}

\begin{figure*}
  \centering
  \includegraphics[width=0.95\linewidth]{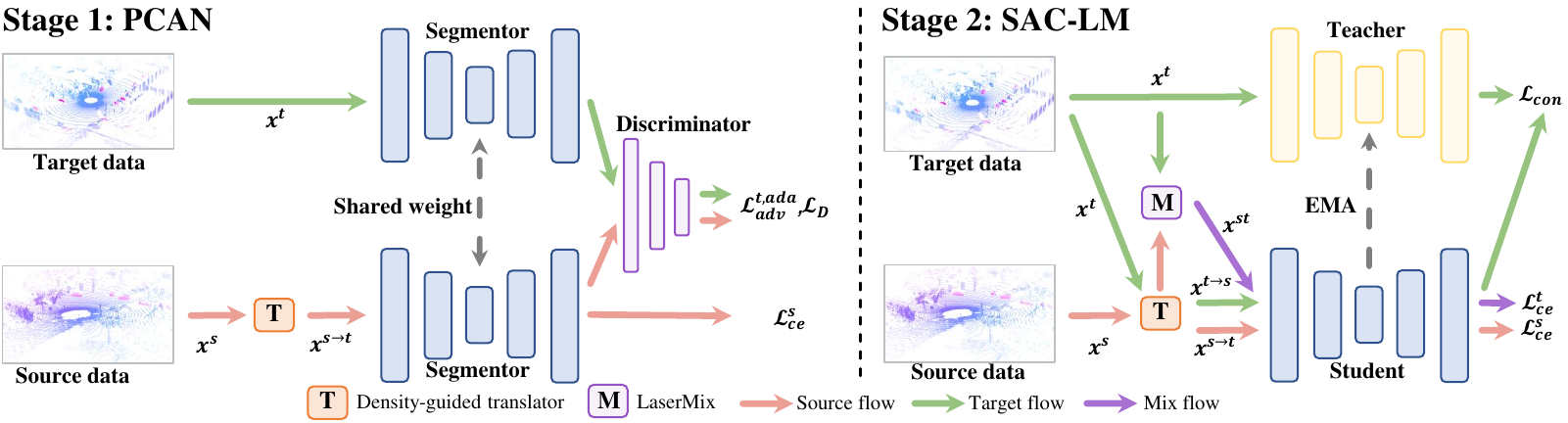}
  \caption{
Overview of our two-stage DGT-ST. We propose DGT to bridge the domain gap at the input level and be integrated into both stages. In stage one, we propose PCAN with a segmentor $G$ and a discriminator $D$. We take the target-like source $x^{s \rightarrow t}$ and raw target data $x^{t}$ as input to perform the category-level adversarial alignment. In stage two, SAC-LM, a teacher-student learning architecture is employed and loads the stage one trained model. We use LaserMix~\cite{kong2023lasermix} to mix two domain scans $x^{s \rightarrow t}$ and $x^{t}$ and obtain the mixed scan $x^{st}$. Finally, the student model is trained by $x^{s \rightarrow t}$ and $x^{st}$. Moreover, we enforce the student model to give consistent predictions on $x^{t}$ and $x^{t \rightarrow s}$.
}
  \label{fig:net_Arch}
\end{figure*}

In the following, we first provide the necessary preliminaries for 3D UDA segmentation (\cref{preliminaries}). Then, we introduce the scan translation strategy DGT in \cref{method:DGT}. After that, we elaborate on the two-stage training pipeline DGT-ST in \cref{method:PCAN,method:LM}, which is also shown in \cref{fig:net_Arch}.


\subsection{Preliminaries} 
\label{preliminaries}
In 3D synthetic-to-real UDA segmentation, we have the source $ \mathcal{X^S} = \{x_i^s\}_{i=1}^{N^s} $ and target $ \mathcal{X^T} = \{x_i^t\}_{i=1}^{N^t} $ dataset of $N^s$ and $N^t$ scans, respectively. $ \mathcal{X^S} $ has point-wise semantic labels $ \mathcal{Y^S} = \{y_i^s\}_{i=1}^{N^s} $, while we lack labels for $ \mathcal{X^T}$. We aim to leverage these datasets to train a segmentation network $G$ that can provide accurate results on $ \mathcal{X^T}$. Due to the domain gap, $G$ trained only on the source data cannot generalize well to the target. Thus, $G$ needs to have both discriminability and transferability. Generally, to ensure discriminability, $G$ is optimized on source by cross-entropy (CE) loss:
\begin{equation}
    \mathcal{L}_{ce}^s = -\frac{1}{N}\sum_{i=1}^{N}\sum_{k=0}^{K} y_i^s \mathrm{log} P_{i, k}^{s},
\end{equation}
where $N$ and $K$ are the number of points and classes in the current training batch, respectively. $y_i^s$ and $P_{i, k}^{s}$ are the ground-truth and probability of the $i$-th point, respectively.

For transferability, the prevalent strategies are adversarial training and self-training. To ensure the features/predictions of $G$ domain-invariant, the adversarial methods adopt a domain discriminator $D$ plus an auxiliary adversarial loss. Here, we show the LS-GAN~\cite{mao2017least} and self-information $S_i^t = -P_i^{t} \mathrm{log} P_i^{t}$ adopted by ADVENT~\cite{vu2019advent}:
\begin{equation}
    \label{adv_loss}
    \mathcal{L}_{adv}^t = -\frac{1}{N^t}\sum_{i=1}^{N^t}||D(S_i^t)-0||_2,
\end{equation}
\begin{equation}
    \label{adv_loss}
    \mathcal{L}_{D} = -\frac{1}{N^s}\sum_{i=1}^{N^s}||D(S_i^s)-0||_2 -\frac{1}{N^t}\sum_{i=1}^{N^t}||D(S_i^t)-1||_2,
\end{equation}
where 0/1 denotes the source/target domain label.

On the other hand, self-training uses the pseudo-label $\hat{y}^t$ to optimize $G$ and gain knowledge from the target domain:
\begin{equation}
    \mathcal{L}_{ce}^t = -\frac{1}{N}\sum_{i=1}^{N}\sum_{k=0}^{K} \hat{y}_i^t \mathrm{log} P_{i, k}^{t}.
\end{equation}
Typically, the mean-teacher framework~\cite{tarvainen2017mean} is adopted to provide accurate and robust $\hat{y}^t$. The student model $G$, parameterized by $\theta^{stu}$, is trained by gradient descent. 
And $\theta^{tea}$, the weights of the teacher model $G^{tea}$, is updated every $t$ iterations with an exponential moving average (EMA) manner:
\begin{equation}
\label{ema_udpate}
    \theta^{tea}_i = \alpha \theta^{tea}_{i-t} + (1- \alpha)\theta^{stu}_i,
\end{equation}
where $i$ denotes the current training iteration and $\alpha$ is a smoothing coefficient that determines the update speed.
Finally, $\hat{y}^t$ is obtained by a confidence strategy:
\begin{equation}
\label{gen_pseudo_label}
    \hat{y}_i^t = 
    \begin{cases}
        \mathop{\arg\max}\limits_{k}p_i^{t, tea}, & \text{max$(p_i^{t, tea}) > Th_p$,} \\
        0,                                 & \text{otherwise,}
    \end{cases}\\
\end{equation}
where $Th_p$ is the threshold to obtain reliable pseudo-labels. 


\subsection{Density-guided translator} 
\label{method:DGT}
The sensor sampling pattern mismatch between the source and target domains is the primary cause of the domain gap. By taking a closer look at the synthetic and real-world scans, as shown in \cref{fig:1}, we observe two significant disparities: (1) The density varies across domains, \ie, the number of beams $D_b$ and the point number per beam $D_p$; (2) The synthetic scan is integral and clean, whereas the real-world data contains a varying amount of noise. To mitigate the domain discrepancy at the input level, \ie, obtain $D_b^s \approx D_b^t$ and $D_p^s \approx D_p^t$, 
we can accomplish this in two ways: 
complete surface~\cite{yi2021complete} or discard points~\cite{xiao2022transfer,li2023adversarially,wu2023virtual}. However, surface completion inevitably increases the computational overhead and may bring in points with inaccurate labels. Thus, we choose to discard points and propose a density-guided translator (DGT), a non-learnable and statistical-based translator.

\begin{figure}[t]
  \centering
  \includegraphics[width=1\columnwidth]{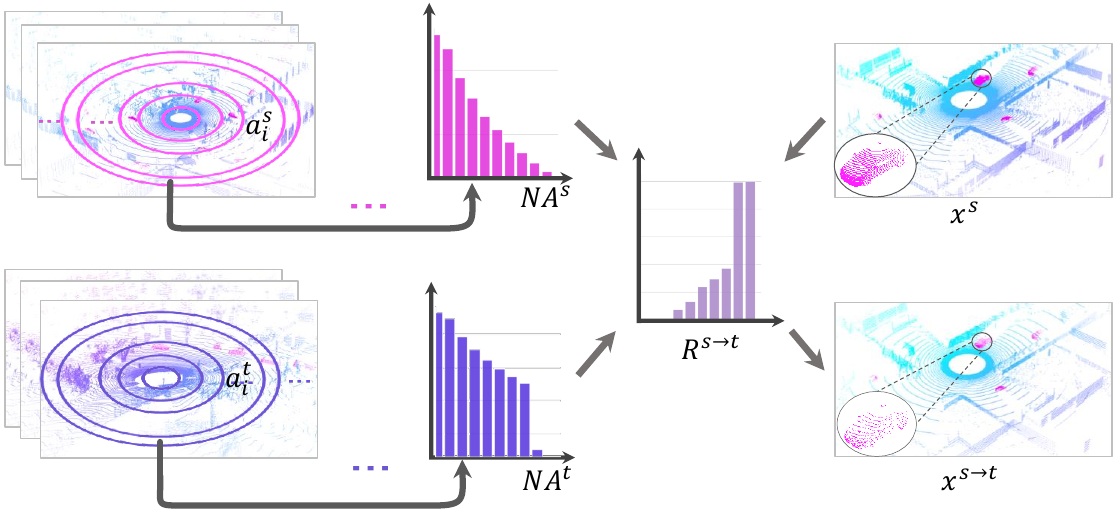}
   \vspace{-0.5cm}
   \caption{Illustration of the density-guided translator (DGT).
   }
   \label{fig:DGT}
\vspace{-0.5cm} 
\end{figure}

Inspired by LiDAR-Distillation~\cite{wei2022lidar}, we use K-means to label beams in each scan and discard beams to make the two domains have a similar number of beams. However, it does not work on two domains with the same number of beams while $D_p$ is still different between domains. $D_p$ exhibits substantial variation according to the distance to the LiDAR center, where $D_p$ in the nearby area is much greater than the farther-away area. Two critical problems arise: \textit{where} and \textit{how many} points to discard? To tackle these issues, we propose a statistical-based random discarding strategy to balance the point number for both domains. As shown in \cref{fig:DGT}, it mainly consists of the following steps:

(1) \textbf{Partition}. For both domains, we evenly partition all points within each scan into $m$ non-overlapping areas $A = \{a_1, a_2,..., a_m\}$ by their distance to the LiDAR center.

(2) \textbf{Calculation}. For each domain, we count the number of points in each area on the entire dataset to obtain $NA^s = \{na_1^s, na_2^s,..., na_m^s\}$ and $NA^t = \{na_1^t, na_2^t,..., na_m^t\}$. The location and number of points that are chosen to be discarded are determined by calculating $R = [r_1, r_2,..., r_m]$, which can be formulated as:
\begin{equation}
\begin{aligned}
    R^{s \rightarrow t} = \left[na_1^t / na_1^s, na_2^t / na_2^s,..., na_m^t / na_m^s \right], \\
    R^{t \rightarrow s} = \left[na_1^s / na_1^t, na_2^s / na_2^t,..., na_m^s / na_m^t \right],
\end{aligned}
\end{equation}
where $s \rightarrow t$ denotes generating a target-like source scan, vice versa. $R$ directly reveals the point number differences within each area for both domains. $R$ will be further clipped to [0, 1], \ie, \texttt{R=np.clip(R, a\_min=0,a\_max=1.)}.



(3) \textbf{Translation}. For simplicity, we give an example of translating a target-like source scan $x^{s \rightarrow t}$. We first count the number of points within each area and obtain $A^{x^s} = \{a_1^{x^s}, a_2^{x^s},..., a_m^{x^s}\}$. Then, the discarding point number $ Del_{a_i}$ of each area is:
\begin{equation}
 Del_{a_i} = 
 \begin{cases}
     a_i^{x^s} * (1 - R^{s \rightarrow t}_i),  & \text{if $R^{s \rightarrow t}_i \le 1$}, \\
     0, & \text{otherwise.}
 \end{cases}
\end{equation}
Finally, to obtain $x^{s \rightarrow t}$, we randomly select $Del_{a_i}$ points within the area where $R^{s \rightarrow t}_i \le 1$ and random noise is additionally added to the $X$ and $Y$ axes to enhance its realism. This strategy can also be employed to acquire $x^{t \rightarrow s}$.


\textbf{Discussion: how about random discarding?} Random discarding does not align with our aim for the following reasons: (1) Points located in farther-away areas already exhibit high sparsity, and discarding these points would impede model discriminability; (2) In certain regions, $D_p^s $/$D_p^t$ may be smaller than $D_p^t$/$D_p^s$, and discarding these points may enlarge the domain gap. We verify this in \cref{ablations}.



\subsection{Category-level adversarial with prototype}
\label{method:PCAN}

Traditional adversarial methods~\cite{tsai2018learning,vu2019advent} adopt a discriminator to bridge the domain gap. However, striving to make the source and target marginal distributions match globally is prone to negative transfer, \ie, well-aligned points mapped to incorrect semantic categories. We resort to prototype and propose PCAN to tackle this issue, as shown in \cref{fig:net_Arch} (left).

Inspired by previous works~\cite{zhang2019category,zhang2021prototypical}, points belonging to the same category tend to cluster together, and the prototype (class centroid) can represent each class in the feature space. Since the source data is fully labeled without noise, we only use source data to calculate the prototype. We define $\lambda_{k}^s$ as the prototype of class $k$, which can be calculated by:
\begin{equation}
    \lambda_{k}^s = \frac{1}{|\mathcal{X}_{k}^S |}\sum_{x_i^s \in \mathcal{X}_{k}^S}G(x_i^s),
\end{equation}
where $\mathcal{X}_{k}^S$ denotes all points whose labels are $k$ in $\mathcal{X}^S$. 

Instead of globally matching the source and target marginal distributions, we propose to perform category-level alignment. Specifically, $G$ should make high-confidence predictions for those well-aligned points, and we can easily obtain their pseudo-labels through~\cref{gen_pseudo_label}. Thus, we treat those points with high confidence as well-aligned points $x_{wa}^t$, and evaluate how well $x_{wa}^t$ are semantically matched based on the similarity distance $\mathcal{M}$ (\eg., cosine similarity) between their feature and the corresponding prototype:
\begin{equation}
 \mathcal{M}(x_{wa, k}^t, \lambda_{k}^s) = 
 \begin{cases}
     1 - \frac{\langle G(x_{wa, k}^t), \lambda_{k}^s \rangle }{  \| G(x_{wa, k}^t) \Vert \cdot \| \lambda_{k}^s\Vert  }, & \text{if $x_i^t \in x_{wa}^t, $}\\
     1, & \text{otherwise.}
 \end{cases}
\end{equation}
$\mathcal{M}$ reveals the alignment degree between the well-aligned points and their corresponding source domain class in the feature space. The more similar $x_{wa, i}^t$ and $ \lambda^s$ are, the smaller the weight is. For the remaining points (\ie, $x^t \notin x_{wa}^t$ in the current batch), we treat them as the unlabel class and calculate their adversarial loss in the traditional manner. Then, we leverage $\mathcal{M}$ to reweight the adversarial loss, adaptively reduce their impact on $x_{wa}^t$, and prevent negative transfer. 

Besides, we adopt a class-wise aggregation strategy to individually calculate each appearing class in $x_{wa}^t$ to reduce interference with the other classes. With the help of the above design, we explicitly incorporate the class information into the adversarial loss. The new category-level adaptive reweight adversarial loss $\mathcal{L}_{adv}^{t,ada}$ can be written as:
\begin{equation}
    \label{rw_adv_loss}
    \mathcal{L}_{adv}^{t,ada}=-\sum_{k=0}^{K}\frac{1}{N_k} \mathcal{M}||D(S_i^t)-0||_2 .
\end{equation}


In this adversarial training stage, we use the translated source scan $x^{s \rightarrow t}$ and target scan $x^t$ to perform the category-level adversarial alignment. The final loss is: 
\begin{equation}
    \label{final_adv_loss}
    \mathcal{L}_{adv}^{total}=\mathcal{L}_{ce}^s(x^{s \rightarrow t}) + \gamma_1\mathcal{L}_{adv}^{t,ada}(x^t),
\end{equation}
where $\gamma_1$ is a balance parameter. Following previous adversarial training works~\cite{tsai2018learning,luo2019taking,vu2019advent,yuan2023prototype}, we fixed it as 0.001.

 \subsection{Source-aware consistency LaserMix}
 \label{method:LM}
 Although the source and target domains have distinct sampling patterns, they are collected in driving scenarios with similar scene layout. Inspired by LaserMix~\cite{kong2023lasermix}, initially proposed for semi-supervised learning (SSL), we conjecture that the spatial positions of objects/background in both domain scans correlate with their respective distributions. Thus, as shown in \cref{fig:net_Arch} (right), we extend it into UDA and propose SAC-LM to mitigate the domain discrepancy.
 


\textbf{LaserMix (LM).} Given two scans $x^s$ and $x^t$ from two domains, we first partition all points from each scan based on their inclination angles and form $n$ non-overlapping areas $A\_M = \{a\_m_1, a\_m_2,..., a\_m_n\}$. Then, we mix $A\_M^s$ and $A\_M^t$ in an intertwining manner and result in two mixed scans $x^{st}_1$ and $x^{st}_2$, which can be formulated as:
\begin{equation}
    \label{mixed_scans}
    \begin{aligned}
    x^{st}_1 = a\_m_1^s \cup a\_m_2^t \cup ... \cup a\_m_{n-1}^s \cup a\_m_n^t, \\
    x^{st}_2 = a\_m_1^t \cup a\_m_2^s \cup ... \cup a\_m_{n-1}^t \cup a\_m_n^s. \\
    \end{aligned}
\end{equation}
Since the target domain is unlabeled, we adopt the teacher model and \cref{gen_pseudo_label} to obtain $\hat{y}^t$ for $x^t$. The semantic labels (\ie, $y^s$ and $\hat{y}^t$) of the two scans are mixed similarly.

\begin{table*}
    \centering
    \resizebox{\linewidth}{!}{
        \large
        \begin{tabular}{@{}l|c|ccccccccccccccccccc|cc@{}}
            \toprule
            Methods                             & \rotatebox{90}{Mech.} & \rotatebox{90}{car} & \rotatebox{90}{bi.cle} & \rotatebox{90}{mt.cle} & \rotatebox{90}{truck} & \rotatebox{90}{oth-v.} & \rotatebox{90}{pers.} & \rotatebox{90}{bi.clst} & \rotatebox{90}{mt.clst} & \rotatebox{90}{road} & \rotatebox{90}{parki.} & \rotatebox{90}{sidew.} & \rotatebox{90}{other-g.} & \rotatebox{90}{build.} & \rotatebox{90}{fence} & \rotatebox{90}{veget.} & \rotatebox{90}{trunk} & \rotatebox{90}{terr.} & \rotatebox{90}{pole} & \rotatebox{90}{traf.} & \rotatebox{90}{mIoU} & \rotatebox{90}{gain} \\
            \midrule
            Source only                         & -                     & 35.9                & 7.5                    & 10.7                   & 0.6                   & 2.9                    & 13.3                  & 44.7                    & \underline{3.4}         & 21.8                 & 6.9                    & 29.6                   & 0.0                      & 34.1                   & 7.4                   & 62.9                   & 26.0                  & 35.5                  & 30.3                 & 14.1                  & 20.4                 & +0.0                 \\
            \midrule
            AdaptSegNet~\cite{tsai2018learning} & A                     & 52.1                & 10.8                   & 11.2                   & 2.6                   & 9.6                    & 15.1                  & 35.9                    & 2.6                     & 62.2                 & 10.4                   & 41.3                   & 0.1                      & 58.1                   & 17.1                  & 68.0                   & 38.4                  & 38.7                  & 35.9                 & 20.4                  & 27.9                 & +7.5                 \\

            CLAN~\cite{luo2019taking}           & A                     & 51.0                & \underline{15.8}       & 16.8                   & 2.2                   & 7.8                    & 18.7                  & 46.8                    & 3.0                     & 68.9                 & 11.1                   & 44.9                   & 0.1                      & 59.6                   & 17.5                  & 71.7                   & 41.1                  & 44.0                  & \underline{37.7}     & 19.8                  & 30.5                 & +10.1                \\

            ADVENT~\cite{vu2019advent}          & A                     & 59.9                & 13.8                   & 14.6                   & \underline{3.0}       & 8.0                    & 17.7                  & 45.8                    & 3.0                     & 67.6                 & \underline{11.3}       & 45.6                   & 0.1                      & \underline{61.7}       & 15.8                  & \underline{72.4}       & \textbf{41.5}         & \underline{47.0}      & 34.5                 & 15.3                  & 30.5                 & +10.1                \\


            FADA~\cite{wang2020classes}         & A                     & 49.9                & 6.7                    & 5.1                    & 2.5                   & \underline{10.0}       & 5.7                   & 26.6                    & 2.3                     & 65.8                 & 10.8                   & 37.8                   & 0.1                      & 60.3                   & \textbf{21.5}         & 60.4                   & 37.2                  & 31.9                  & 35.4                 & 17.4                  & 25.6                 & +5.2                 \\

            MRNet~\cite{zheng2020unsupervised}  & A                     & 49.5                & 11.0                   & 12.2                   & 2.2                   & 8.6                    & 16.0                  & 46.4                    & 2.7                     & 60.0                 & 10.5                   & 41.9                   & 0.1                      & 55.1                   & 16.5                  & 68.1                   & 38.0                  & 40.7                  & 36.5                 & \underline{20.8}      & 28.3                 & +7.9                 \\

            PMAN~\cite{yuan2023prototype}                                & A                     & \underline{71.0}    & 14.9                   & \underline{24.8}       & 1.6                   & 6.6                    & \underline{23.6}      & \underline{61.1}        & \textbf{5.5}            & \underline{75.3}     & 10.5                   & \textbf{54.1}          & 0.1                      & 47.9                   & 17.4                  & 69.6                   & 38.6                  & \textbf{61.5}         & 37.0                 & 18.6                  & \underline{33.7}                 & +13.3                \\
            PCAN (Ours)                     & A                     & \textbf{85.0}       & \textbf{17.5}          & \textbf{27.4}          & \textbf{10.4}         & \textbf{11.9}          & \textbf{27.5}         & \textbf{63.7}           & 2.6                     & \textbf{78.1}        & \textbf{13.5}          & \underline{50.1}       & 0.1                      & \textbf{68.5}          & \underline{20.0}      & \textbf{76.2}          & \underline{41.3}      & 45.7                  & \textbf{41.0}        & \textbf{21.8}         & \textbf{37.0}                 & +16.6                \\

            \midrule

            CoSMix~\cite{saltori2022cosmix}     & S                     & 56.4                & \underline{10.2}       & 20.8                   & 2.1                   & \textbf{13.0}          & 25.6                  & 41.3                    & 2.2                     & 67.4                 & 8.2                    & 43.4                   & 0.0                      & \underline{57.9}       & \textbf{12.2}         & 68.4                   & 44.8                  & 35.0                  & \underline{42.1}     & \underline{17.0}      & 29.9                 & +9.5                 \\

            PolarMix~\cite{xiao2022polarmix}    & S                     & -                   & -                      & -                      & -                     & -                      & -                     & -                       & -                       & -                    & -                      & -                      & -                        & -                      & -                     & -                      & -                     & -                     & -                    & -                     & 31.0                 & +10.6                \\

            LaserMix~\cite{kong2023lasermix}  & S                     & \underline{90.3}    & 7.8                    & \underline{37.2}       & \underline{2.3}       & 2.4                    & \underline{40.6}      & \underline{49.1}        & \textbf{5.1}            & \underline{80.5}     & \underline{9.9}        & \underline{57.4}       & 0.0                      & 57.6                   & 3.4                   & \underline{77.6}       & \underline{46.6}      & \textbf{60.1}         & 42.0                 & 13.6                  & \underline{36.0}                 & +15.6                \\

            DGT-ST (Ours)                     & S                     & \textbf{92.9}       & \textbf{17.3}          & \textbf{43.4}          & \textbf{15.0}         & \underline{6.1}        & \textbf{49.2}         & \textbf{54.2}           & \underline{4.2}         & \textbf{86.4}        & \textbf{19.1}          & \textbf{62.3}          & 0.0                      & \textbf{78.2}          & \underline{9.2}       & \textbf{83.3}          & \textbf{56.0}         & \underline{59.1}      & \textbf{51.2}        & \textbf{32.3}         & \textbf{43.1}                 & +  22.7              \\

            \bottomrule
        \end{tabular}
    }
    \caption{Comparison results of SynLiDAR $\rightarrow$ semanticKITTI adaptation in terms of mIoU. A/S denotes adversarial training/self-training.}
    \label{tab:Syn2Sk_XYZ_tab}
\end{table*}

\textbf{Source-aware consistency (SAC) regularization.} In \cref{fig:1} bottom, in certain areas, the target domain point number is larger than the source. We can easily generate the source-like target scan by DGT. When $G$ is domain-invariant, $G$ should give similar predictions on raw target input $x^t$ and its source-like input $x^{t \rightarrow s}$. We minimize the Kullback–Leibler divergence between $P^{tea}(x^t)$ and $P(x^{t \rightarrow s})$:
\begin{equation}
    \mathcal{L}_{sac} = -\frac{1}{N}\sum_{i=1}^{N}P(x^{t \rightarrow s})\frac{P(x^{t \rightarrow s})}{P^{tea}(x^t)}.
\end{equation}
$\mathcal{L}_{sac}$ enforces $G$ to give consistent predictions for a target input under two views in the mean-teacher framework. It enables the segmentor to acquire source knowledge from the target data, thereby enhancing the capability of domain-invariant feature extraction and reducing the domain gap.


We use PCAN to provide a well-initialized model for generating high-quality pseudo-labels in this stage. We use the translated source scan $x^{s \rightarrow t}$ and the mixed scans $x^{st}_1$ to perform self-training. Since DGT is used in both stages, we name this overall pipeline as DGT-ST, and the final loss is: 
\begin{equation}
    \label{final_st_loss}
    \mathcal{L}_{st}=\mathcal{L}_{ce}^s(x^{s \rightarrow t}) + \mathcal{L}_{ce}^t(x^{st}_1)+\gamma_2\mathcal{L}_{sac},
\end{equation}
where $\gamma_2$ is a balance parameter, and we fixed it as 0.001.

\section{Experiments}
\label{sec:exps}

\subsection{Setup}
\textbf{Datasets.} We perform two synthetic-to-real UDA tasks.

\textit{SynLiDAR}~\cite{xiao2022transfer} is a recently published synthetic dataset generated by a LiDAR simulator identical to the Velodyne HDL-64E with a 100-meter working range and contains a variety of realistic virtual scenarios. Following the official recommendation, we use all subdataset, which contains 13 sequences of about 19840 scans. 

\textit{SemanticKITTI}~\cite{behley2019semantickitti} is the most prevalent real-world dataset for evaluating the large-scale 3D segmentation method, collected in Germany by a Velodyne HDL-64E LiDAR sensor. Following ~\cite{xiao2022transfer,yuan2023prototype,saltori2022cosmix,xiao2022polarmix}, we choose sequences 00-10 for training (19130 scans) except sequence 08 (4071 scans) for validation. 

\textit{SemanticPOSS}~\cite{pan2020semanticposs} is a real-world dataset, collected in Peking University by a Pandora 40-line LiDAR sensor. Following ~\cite{xiao2022transfer,xiao2022polarmix,yuan2023prototype}, sequence 03 (500 scans) is used for validation and the rest (2488 scans) for training.

\textbf{Evaluation protocol.} The SynLiDAR provides the mapping details to pair with SynLiDAR$\rightarrow$SemanticKITTI (Syn$\rightarrow$Sk) and SynLiDAR$\rightarrow$SemanticPOSS (Syn$\rightarrow$Sp). We can fairly compare DGT-ST with other methods. Following~\cite{saltori2022cosmix,xiao2022polarmix,yuan2023prototype}, the segmentation performance is reported using the mean Intersection over Union (mIoU, as $\%$) metric.



\textbf{Implementation.} All experiments are implemented in PyTorch~\cite{paszke2019pytorch} and MinkowskiEngine~\cite{choy20194d} on a single NVIDIA RTX 3090 GPU. For a fair comparison with other methods, we adopt MinkUNet34 as the segmentation network and set the voxel size as 0.05$m$. For PCAN, the discriminator consists of 5 sparse convolution layers with kernel size 4 and stride 2, where the channel number is \{32, 64, 64, 128, 1\}. A Leaky-ReLU activation layer follows each convolution layer except the last. To obtain the same size as the input, we upsample the discriminator output by interpolation, \ie, \texttt{ME.MinkowskiInterpolation()}. We adopt the Adam~\cite{kingma2014adam} optimizer with the initial learning rate of 2.5e-4 and 1e-4 respectively for the segmentation network and discriminator and is decayed by a poly learning rate policy with power of 0.9. The batch size is set to 2 and the max training iteration of all experiments is set as 100K. The input feature of all methods is XYZ coordinates. Following CoSMix~\cite{saltori2022cosmix}, we set $Th_p$, $\alpha$ and $t$ as 0.9, 0.99 and 100, respectively. 


\subsection{Comparisons with previous methods} We comprehensively compare our proposed method with the recent state-of-the-art approaches. These methods could be divided into two groups: (1) adversarial training methods, including AdaptSeg~\cite{tsai2018learning}, CLAN~\cite{luo2019taking}, ADVENT~\cite{vu2019advent}, FADA~\cite{wang2020classes}, MRNet~\cite{zheng2020unsupervised}, and PMAN~\cite{yuan2023prototype}; (2) self-training methods, including CosMix~\cite{saltori2022cosmix}, PolarMix~\cite{xiao2022polarmix} and LaserMix (LM)~\cite{kong2023lasermix}. Source only denotes the model trained on the source without adaptation. The mechanism ``A" and ``S" denotes adversarial training and self-training, respectively. 
%

\textbf{SynLiDAR$\rightarrow$SemanticKITTI.} In ~\cref{tab:Syn2Sk_XYZ_tab}, we show the comparison results. PCAN and DGT-ST significantly outperform the other methods, yielding an accuracy of 37.0$\%$ and 43.1$\%$ in mIoU. Compared with the non-adapted source only, PCAN and DGT-ST offer mIoU gains of 16.6$\%$ and 22.7$\%$. PCAN outperforms the second-best adversarial method PMAN by 3.3$\%$. Among these adversarial training methods, PCAN and PMAN are two specifically designed category-level adversarial networks for 3D UDA segmentation. Their results show that class-level adversarial alignment is effective for this task. For self-training methods, LM significantly outperforms other methods, which means that the spatial prior helps bridge the domain gap. Unlike CoSMix, which requires hyperparameters to select the number of categories to mix and then select and paste the reliable points to boost the performance of rare classes (\eg, oth-v.), LM is more convenient. Among all the 19 classes, DGT-ST obtains the best results in 14 classes with significant improvements. 



\begin{table*}
    \centering
    \resizebox{\linewidth}{!}{
        \small
        \begin{tabular}{@{}l|c|ccccccccccccc|cc@{}}
            \toprule

            Methods                               & Mech. & bi.clst          & car              & trunk            & veget.           & traf.            & pole             & garb.            & build.           & cone.            & fence            & bi.cle           & ground           & pers.            & mIoU             & gain  \\

            \midrule

            Source only                           & -     & 47.2             & 43.6             & 37.8             & 70.3             & 11.1             & 33.8             & 19.5             & 67.9             & 11.2             & 19.9             & 9.6              & \underline{77.9} & 47.8             & 38.3             & +0.0  \\
            \midrule
            AdaptSegNet ~\cite{tsai2018learning}  & A     & 43.9             & 48.2             & 39.0             & 69.6             & 15.5             & 33.6             & 21.3             & 64.3             & 12.7             & 25.0             & 11.6             & 76.0             & 49.9             & 39.3             & +1.0  \\

            CLAN    ~\cite{luo2019taking}         & A     & 43.9             & 46.6             & \underline{41.3} & 71.0             & 15.1             & 34.3             & 20.4             & 69.6             & 9.5              & 23.2             & 12.0             & 75.1             & 51.3             & 39.5             & +1.2  \\

            ADVENT~\cite{vu2019advent}            & A     & 44.6             & 47.6             & 40.3             & 71.2             & 15.6             & 35.6             & \underline{22.0} & 68.4             & 10.6             & \underline{25.9} & 10.4             & 76.7             & 52.3             & 40.1             & +1.8  \\


            FADA ~\cite{wang2020classes}          & A     & 39.6             & 41.2             & 38.8             & 69.2             & 16.3             & 32.1             & 18.1             & 67.9             & 11.5             & 22.0             & \underline{13.0} & 71.4             & 47.9             & 37.6             & -0.7  \\

            MRNet   ~\cite{zheng2020unsupervised} & A     & 43.5             & 47.2             & 39.1             & 70.4             & 15.5             & 32.8             & \underline{22.0} & 66.1             & \underline{13.2} & 24.2             & 11.2             & 76.8             & 50.0             & 39.4             & +1.1  \\

            PMAN~\cite{yuan2023prototype}                                 & A     & \textbf{52.6}    & \underline{61.5} & \textbf{46.8}    & \textbf{75.1}    & \underline{18.8} & \textbf{36.5}    & 21.4             & \underline{74.7}    & \textbf{18.3}    & 25.8             & \textbf{37.5}    & 73.7             & \textbf{61.9}    & \textbf{46.5}    & +8.2  \\

            PCAN (Ours)                              & A     & \underline{48.6} & \textbf{62.1}    & 37.5             & \underline{74.0} & \textbf{23.9}    & \underline{31.4} & \textbf{22.2}    & \textbf{76.9} & 6.5              & \textbf{41.9}    & 11.9             & \textbf{79.1}    & \underline{61.2} & \underline{44.4} & +6.1  \\

            \midrule
            CoSMix~\cite{saltori2022cosmix}       & S     & \underline{53.6} & 47.6             & 44.8             & \textbf{75.1}    & 16.8             & \underline{37.9} & 25.3             & \underline{72.7} & \textbf{19.9}    & \underline{39.7} & \textbf{10.8}    & \textbf{80.0}    & 56.5             & 44.6             & +6.3  \\
            PolarMix~\cite{xiao2022polarmix}      & S     & -                & -                & -                & -                & -                & -                & -                & -                & -                & -                & -                & -                & -                & 30.4             & -8.3  \\
            LaserMix~\cite{kong2023lasermix}                            & S     & \textbf{58.4}    & \underline{61.3} & \textbf{47.7}    & 69.0             & \underline{21.9} & \textbf{39.5}    & \underline{30.9} & 61.0             & \underline{16.1} & 36.5             & 7.1              & \underline{79.5} & \underline{62.6} & \underline{45.5} & +7.2  \\
            DGT-ST (Ours)                              & S     & 55.1             & \textbf{70.7}    & \underline{46.1} & \underline{74.2} & \textbf{30.1}    & 36.3             & \textbf{44.1}    & \textbf{81.0}    & 4.3             & \textbf{62.8}    & \underline{10.3} & 78.5             & \textbf{67.2}    & \textbf{50.8}    & +12.5 \\

            \bottomrule
        \end{tabular}
    }
    \caption{Comparison results of SynLiDAR $\rightarrow$ semanticPOSS adaptation in terms of mIoU. A/S denotes adversarial training/self-training.}
    \label{tab:Syn2Sp_XYZ_tab}
\end{table*}

\begin{table}[t]
    \centering
    \resizebox{1\linewidth}{!}
    {
        \begin{tabular}{@{}l|c|cccc|cc@{}}
            \toprule
              & Baseline                       & PCAN       & XY-noise   & Random $D_p$ & Density $D_p$ & mIoU & gain  \\ 
            \midrule
            0 & \multirow{5}{*}{ADVENT}      &            &            &              &               & 30.5 & +0.0  \\
            1 &                              & \Checkmark &            &              &               & 35.4 & +4.9  \\
            2 &                              & \Checkmark & \Checkmark &              &               & 35.8 & +5.3  \\
            3 &                              & \Checkmark & \Checkmark & \Checkmark   &               & 34.9 & +4.4  \\
            4 &                              & \Checkmark & \Checkmark &              & \Checkmark    & \textbf{37.0} & +\textbf{6.5}  \\
            \midrule
              &                              & LM        & DGT        & SAC    & PCAN model         &      &       \\ 
            \midrule
            5 & \multirow{5}{*}{Source only} &            &            &              &               & 20.4 & +0.0  \\
            6 &                              & \Checkmark &            &              &               & 36.0 & +15.6 \\
            7 &                              & \Checkmark & \Checkmark &              &               & 37.5 & +17.1 \\
            8 &                              & \Checkmark & \Checkmark & \Checkmark   &               & 38.7 & +18.3 \\
            9 &                             & \Checkmark &            &              &  \Checkmark   & 39.8 & +19.4 \\
            10 &                              & \Checkmark & \Checkmark & \Checkmark   & \Checkmark    & \textbf{43.1} & +\textbf{22.7} \\

            \bottomrule
        \end{tabular}
    }
    \caption{Ablation of each component in DGT-ST on Syn $\rightarrow$ Sk. The whole training consists of adversarial training (upper) and self-training (bottom). The PCAN model represents the initialization model of the self-training provided by PCAN.}
    \label{tab:ab_dgtst}
\end{table}



\textbf{SynLiDAR$\rightarrow$SemanticPOSS.} We present the results in~\cref{tab:Syn2Sp_XYZ_tab}. PCAN achieves competitive performance in adversarial-based methods, yielding an accuracy of 44.4$\%$ in mIoU and outperforming the source-only model by 6.1$\%$. Compared with PMAN, which utilizes a multi-task network and is trained with specially designed losses, PCAN is much simpler. DGT-ST achieves an mIoU score 50.8 $\%$ and outperforms the source only model by 12.5$\%$. DGT-ST outperforms all compared methods by a large margin and achieves the top performance in 6 out of the 13 categories.


\subsection{Ablation Studies}
\label{ablations}

We report DGT-ST with certain components ablated here. Since the SynLiDAR and semanticKITTI are collected by 64-beam LiDAR, discarding beams does not work on Syn $\rightarrow$ Sk. Thus, we validate the effectiveness of discarding beams and points of DGT on Syn $\rightarrow$ Sp. For others, we only present the results on Syn $\rightarrow$ Sk due to 
its much larger validation set than semanticPOSS and limited space.

\textbf{Density-guided translator.} To validate the effectiveness of DGT, we conduct experiments in adversarial training and self-training in~\cref{tab:ab_dgtst,tab:ab_dgt_Syn2Sp}. As shown in \cref{tab:ab_dgtst}, with DGT, we respectively obtain +1.6$\%$ (rows 1 and 4) mIoU gain in PCAN, and +1.5$\%$ (rows 6 and 7) mIoU gain in LM, on Syn $\rightarrow$ Sk task. Specifically, DGT consists of injecting noise on the X and Y axes (XY-noise), discarding beams $D_b$ and discarding points $D_p$. Comparing rows 1 and 2, we get +0.4$\%$ mIoU increase by adding XY-noise. Discarding points based on the point density (Density $D_p$) brings +1.2$\%$ (rows 2 and 4) mIoU improvement. However, randomly discarding points reduces the performance, which drops the mIoU by 0.9$\%$ (rows 2 and 3). These results confirm the discussion part in \cref{method:DGT}. In \cref{tab:ab_dgt_Syn2Sp}, DGT also brings +2.1$\%$ (rows 1 and 3) and +2.2$\%$ (rows 5 and 7) mIoU gains on Syn $\rightarrow$ Sp task. Besides, when the source and target domain are collected by sensors with different beams, DGT will discard both beams and points. We validate each of them on Syn $\rightarrow$ Sp. In \cref{tab:ab_dgt_Syn2Sp}, compared to PCAN and LM, we respectively obtain +1.3$\%$ (rows 1 and 2) and +1.2$\%$ (rows 5 and 6) mIoU gain by discarding beams. Discarding points further improves the performance, which brings +0.8$\%$ and +1.0$\%$ mIoU gain in PCAN and LM, respectively. 

\begin{figure*}
  \centering
  \includegraphics[width=1\linewidth]{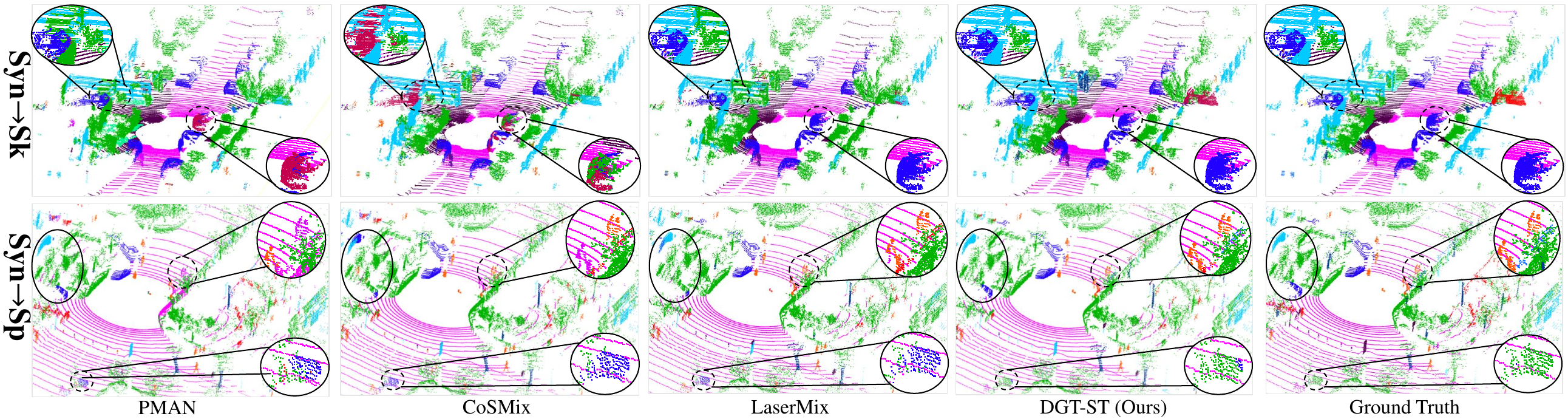}
  \caption{
Visual results of UDA segmentation for SynLiDAR $\rightarrow$ SemanticKITTI and SynLiDAR $\rightarrow$ SemanticPOSS tasks. Black circles highlight some regions of interest. Best viewed in color.
}
  \label{fig:seg_com}
\vspace{-0.4cm} 
\end{figure*}

We further verify the effectiveness of DGT in CoSMix. In \cref{tab:DGTST_CoSMix}, DGT brings +0.6$\%$ (rows 1 and 2) mIoU gain. Notably, CoSMix is already equipped with some data augmentation strategies, including random subsample and scan rotation. DGT still improves its performance. All these experimental results prove that DGT can boost the performance of the prevalent 3D UDA segmentation methods. 


\begin{table}[t]
    \centering
    \resizebox{0.95\linewidth}{!}
    {
        \begin{tabular}{@{}l|c|ccc|cc@{}}
            \toprule
              & Baseline                       & PCAN       & Discard beams & Discard points & mIoU & gain \\ 
            \midrule
            0 & \multirow{4}{*}{ADVENT}      &            &               &                & 40.1 & +0.0 \\
            1 &                              & \Checkmark &               &                & 42.3 & +2.2 \\
            2 &                              & \Checkmark & \Checkmark    &                & 43.6 & +3.5 \\
            3 &                              & \Checkmark & \Checkmark    & \Checkmark     & \textbf{44.4} & +\textbf{4.3} \\
            \midrule
              &                              & LM        & Discard beams & Discard points &      &      \\ 
            \midrule
            4 & \multirow{4}{*}{source only} &            &               &                & 38.3 & +0.0 \\
            5 &                              & \Checkmark &               &                & 45.5 & +7.2 \\
            6 &                              & \Checkmark & \Checkmark    &                & 46.7 & +8.4 \\
            7 &                              & \Checkmark & \Checkmark    & \Checkmark     & \textbf{47.7} & +\textbf{9.4} \\
            \bottomrule
        \end{tabular}
    }
    \caption{Ablations of PCAN and LaserMix on Syn $\rightarrow$ Sp. In addition to XY-noise, DGT needs to discard beams and points, we ablate them in adversarial training and self-training, respectively.}
    \label{tab:ab_dgt_Syn2Sp}
\end{table}

\textbf{PCAN.} We investigate the effectiveness of using the prototype to perform category-level adversarial alignment. In~\cref{tab:ab_dgtst}, the baseline method ADVENT only gives 30.5$\%$ mIoU score on the target domain (row 0). PCAN brings +4.9$\%$ mIoU gain on Syn $\rightarrow$ Sk task (rows 0 and 1). Its effectiveness is also proved in \cref{tab:ab_dgt_Syn2Sp}, where PCAN brings +2.2$\%$ mIoU gain on Syn $\rightarrow$ Sp task (rows 0 and 1).

\textbf{LaserMix.} Although LaserMix (LM) is proposed for 3D SSL, we extend it to 3D UDA segmentation and shows encouraging performance. LM brings +15.6$\%$ (rows 5 and 6 in~\cref{tab:ab_dgtst}) and +7.2$\%$ (rows 4 and 5 in~\cref{tab:ab_dgt_Syn2Sp}) mIoU gain on Syn $\rightarrow$ Sk and Syn $\rightarrow$ Sp, respectively. Compared with CoSMix (results in ~\cref{tab:Syn2Sk_XYZ_tab,tab:Syn2Sp_XYZ_tab}), it respectively outperforms CoSMix by 6.1$\%$ and 0.9$\%$ on two UDA tasks.

\textbf{Source-aware consistency (SAC) regularization.} We aim to use DGT to generate source-like target scans to bridge the domain gap. SAC is employed to give consistent predictions about a target scan with and without DGT. In~\cref{tab:ab_dgtst,tab:DGTST_CoSMix}, its effectiveness is proved by the fact that adding this term offers +1.2$\%$ (rows 7 and 8 in \cref{tab:ab_dgtst}) and +0.9$\%$ (rows 3 and 4 in \cref{tab:DGTST_CoSMix}) mIoU gains for LM and CosMix on Syn $\rightarrow$ Sk task, respectively. These results also validate the effectiveness of our method in the single-stage setting.

\textbf{PCAN pretrained model.} In~\cref{tab:ab_dgtst,tab:DGTST_CoSMix}, we conduct experiments with and without PCAN pretrained model for LM and CoSMix. The pretrained model brings +3.8$\%$ (rows 6 and 9 in \cref{tab:ab_dgtst}) and +8.6$\%$ (rows 2 and 3 in \cref{tab:DGTST_CoSMix}) mIoU gains for them, respectively. It also brings +4.4$\%$ (rows 8 and 10 in \cref{tab:ab_dgtst}) mIoU gain for our final model. These results are consistent with 2D counterparts, \ie, the well-initialized model helps the self-training methods.

\begin{table}[t]
    \centering
    \resizebox{\linewidth}{!}
    {
        \begin{tabular}{@{}l|l|cc@{}}
            \toprule
              & Method                                                                 & mIoU & gain  \\
            \midrule
            0 & Source only                                                            & 20.4 & +0.0  \\
            1 & CoSMix~\cite{saltori2022cosmix}                                        & 29.9 & +9.5  \\
            2 & CoSMix~\cite{saltori2022cosmix} + DGT                                  & 30.5 & +10.1 \\
            3 & CoSMix~\cite{saltori2022cosmix} + DGT + PCAN model            & 39.1 & +18.7 \\
            4 & CoSMix~\cite{saltori2022cosmix} + DGT + PCAN model+ SAC & \textbf{40.0} & +\textbf{19.6} \\
            \bottomrule
        \end{tabular}
    }
    \caption{Validation of the effectiveness of each proposed component with CoSMix~\cite{saltori2022cosmix} on Syn $\rightarrow$ Sk.}
    \label{tab:DGTST_CoSMix}
\end{table}


\begin{table}[t]
    \centering
    \resizebox{0.7\linewidth}{!}
    {
        \begin{tabular}{@{}c|ccccc@{}}
            \toprule
            $\gamma_2$ & 0    & 0.1  & 0.01 & 0.001 & 0.0001 \\
            \midrule
            mIoU     & 37.5 & 37.4 & 38.3 & \textbf{38.7}  & 38.6   \\

            \bottomrule
        \end{tabular}
    }
    \caption{The effect of the hyper-parameter $\gamma_2$ in the second stage of DGT-ST on Syn $\rightarrow$ Sk.}
    \label{tab:praram_gamma_2}
    \vspace{-0.4cm} 
\end{table}

\textbf{Parameter sensitivity of $\gamma_2$.} 
$\gamma_2$ is a hyper-parameter in \cref{final_st_loss} to balance the SAC loss and the other two CE losses. A larger $\gamma_2$ would force the student model to pay more attention to make consistent predictions with the teacher model, thereby influencing the student to learn the target domain knowledge. Thus, we conduct experiments to observe the impact of changing this trade-off value, and the results are shown in \cref{tab:praram_gamma_2}. 
The final result is stable when $\gamma_2$ is smaller than 0.1. A proper choice of $\gamma_2$ is between 0.001 and 0.0001.

\subsection{Qualitative Results} 

\cref{fig:seg_com} presents a visual comparison of DGT-ST against the previous two lines of methods, including adversarial training (PMAN) and self-training (CoSMix and LM). From the first row , we can see that PMAN and CoSMix incorrectly classify the car as tree or road while DGT-ST identifies them precisely. DGT-ST gives more accurate results (top black circle) than LM. In the bottom row, DGT-ST can not only accurately classify the person walking on the road, but also identify the plants that are easily misclassified as cars (bottom black circle). Thus, we can conclude that DGT-ST can extract more discriminative features than the previous works.

\section{Conclusions}
\label{sec:conclusion}

In this paper, we present a LiDAR scan translation strategy DGT and a two-stage training pipeline DGT-ST for 3D synthetic-to-real UDA segmentation. DGT leverages the point density to discard points of each scan, thereby alleviating the domain gap at the input level. It can be integrated into the prevalent UDA methods and boost their final performance. In the first stage of DGT-ST, we propose PCAN to provide a well-initialized pretrained model for self-training. It is a category-level adversarial network and uses prototypes to prevent negative transfer. By leveraging DGT and PCAN, in the second stage of DGT-ST, we use LaserMix to construct an intermediate domain and propose SAC-LM to perform self-training. A source-aware consistent loss is proposed to empower the segmentor to learn the source knowledge from the target data, thereby mitigating the domain discrepancy further. Extensive experiments on two prevalent synthetic-to-real tasks demonstrate the superiority of DGT-ST, which outperforms the previous approaches by a large margin.

{
    \small
    \bibliographystyle{ieeenat_fullname}
    \bibliography{main}
}

\clearpage
\setcounter{page}{1}
\setcounter{section}{0}
\setcounter{figure}{0}
\setcounter{table}{0}
\maketitlesupplementary

In the following sections, we first compare two versions of the noise of DGT in \cref{sup:xy_and_xyz_noise}. Then, we present the hyper-parameter sensitivity analysis in \cref{sup:param_sen}. After that, we present the t-SNE feature visualization of different methods in \cref{sup:tsne_vis}. Finally, we show more quantitative results for a better comparison against previous methods in \cref{sup:more_vis_com}.

\section{Noise injection in DGT}
\label{sup:xy_and_xyz_noise}

\begin{figure}[t]
    \centering
    \includegraphics[width=1\linewidth]{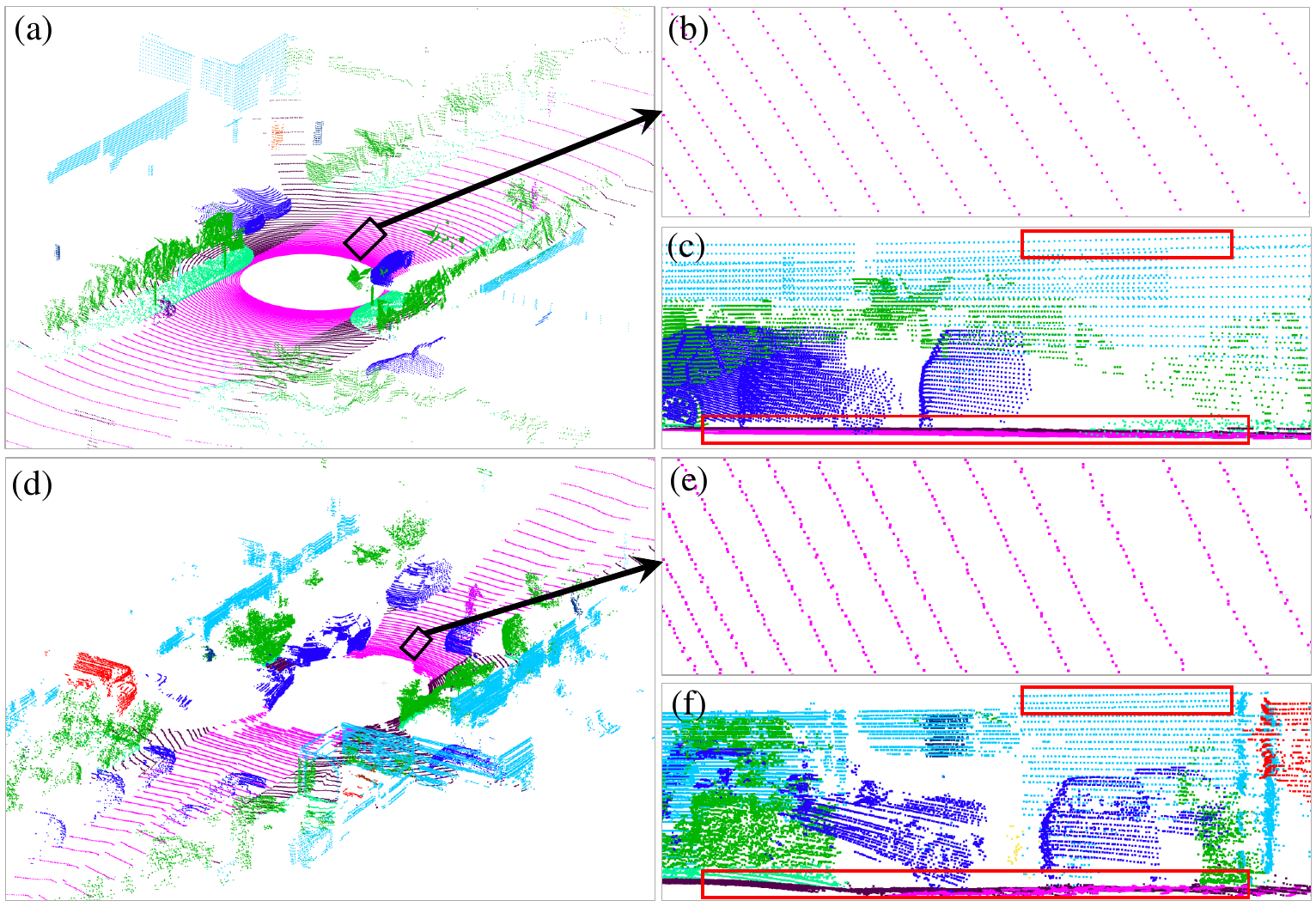}
    \caption{
    Comparison of synthetic and real-world scans. (a) and (d) show one scan of SynLiDAR and SemanticKITTI, respectively. (b) and (e) are zoomed-in visualizations of the road in the black box shown in (a) and (d). (c) and (f) are side-view visualizations of part of (a) and (d). The red boxes in (c) and (f) highlight that the points of synthetic and real-world scans do not exhibit significant shifts along the Z-axis.
    }
    \label{fig:frontView}
\end{figure}

As depicted in \cref{fig:frontView}(e) and \cref{fig:frontView}(f), points of the real-world scan show noticeable shifts in X and Y directions, and the points in the synthetic scan are integral and clean (\cref{fig:frontView}(b) and \cref{fig:frontView}(c)). However, the points in the red box show no significant shifts along the Z-axis in either synthetic or real-world scans. Thus, as stated in the main body of this paper, we do not inject noise on the Z-axis and only add random noise to the X and Y axes of the synthetic scan to enhance its realism. Moreover, we use PCAN and conduct experiments with two versions of DGT, \ie, inject noise on the X and Y axes (XY-noise) and inject noise on the X, Y, and Z axes (XYZ-noise). As shown in \cref{tab:DGT_xyz_shift}, in comparison with XY-noise, XYZ-noise drops mIoU by 1.4$\%$ on SynLiDAR $\rightarrow$ SemanticKITTI (Syn $\rightarrow$ Sk).

\section{Parameter sensitivity}
\label{sup:param_sen}

$\boldsymbol{t}$. $t$ is a hyper-parameter to control the update frequency of the teacher model. The larger $t$ is, the more stable the teacher model is. In this study, we use LaserMix, fix $\alpha$=0.99, and experiment with different $t$ on Syn $\rightarrow$ Sk. As shown in \cref{tab:praram_t}, we get the best performance when $t$=100. A proper choice of $t$ is between 100 and 200. 

$\boldsymbol{\alpha}$. $\alpha$ is a hyper-parameter to control the update speed of the teacher model. A smaller $\alpha$ would render the training unstable, and a larger $\alpha$ would stabilize the model training but impede the student model from acquiring new target knowledge effectively. Here, we use LaserMix, fix $t$=100, and experiment with different $\alpha$ on Syn $\rightarrow$ Sk. As shown in \cref{tab:praram_alpha}, a proper choice of $\alpha$ is between 0.99 and 0.999. 

$\boldsymbol{Th_{p}}$. $Th_{p}$ is the confidence threshold to select the pseudo labels. On the one hand, a smaller $Th_{p}$ would yield many points with pseudo labels, but their accuracy cannot be guaranteed. On the other hand, a larger $Th_{p}$ will filter out many incorrect pseudo-labeled points, but it is also possible to filter out correctly predicted points with smaller confidence. In this study, we experiment with different $Th_{p}$ in our DGT-ST on Syn $\rightarrow$ Sk. We present the results in \cref{tab:praram_thp_sac}, among which we got the best performance when $Th_p$=0.4 and $Th_p$=0.5. However, for a fair comparison with CoSMix, we do not finetune this parameter and use $Th_p$=0.9 in the main body of this paper. Moreover, the final performance of DGT-ST is not sensitive to $Th_p$, and a proper choice of $Th_p$ is between 0.4 and 0.7.

\begin{table}[t]
    \centering
    \resizebox{0.9\linewidth}{!}
    {
        \begin{tabular}{@{}c|cc@{}}
            \toprule
            PCAN & DGT with XY-noise  & DGT with XYZ-noise   \\
            \midrule
            mIoU & \textbf{37.0} & 35.6   \\
            \bottomrule
        \end{tabular}
    }
    \caption{Comparison results of injecting noise on X and Y axes (XY-noise) and injecting noise on X, Y, and Z axes (XYZ-noise) in DGT on Syn $\rightarrow$ Sk.}
    \label{tab:DGT_xyz_shift}
\end{table}

\begin{table}[t]
    \centering
    \resizebox{0.7\linewidth}{!}
    {
        \begin{tabular}{@{}c|ccccc@{}}
            \toprule
            $t$ & 1    & 100 & 200  & 300  & 400  \\
            \midrule
            mIoU     & 32.7 & \textbf{36.0}  & 35.9 & 35.7 & 35.3 \\

            \bottomrule
        \end{tabular}
    }
    \caption{Effect of $t$ in the mean-teacher framework on Syn $\rightarrow$ Sk.}
    \label{tab:praram_t}
\end{table}

\begin{table}[t]
    \centering
    \resizebox{0.7\linewidth}{!}
    {
        \begin{tabular}{@{}c|cccc@{}}
            \toprule
            $\alpha$ & 0.9  & 0.99 & 0.999 & 0.9999 \\
            \midrule
            mIoU     & 30.1 & \textbf{36.0} & 34.6  & 33.1   \\

            \bottomrule
        \end{tabular}
    }
    \caption{Effect of $\alpha$ in the mean-teacher framework on Syn $\rightarrow$ Sk.}
    \label{tab:praram_alpha}
\end{table}

\begin{table*}[t]
    \centering
    \resizebox{0.75\linewidth}{!}
    {
        \begin{tabular}{@{}c|ccccccccccc@{}}
            \toprule
            $Th_{p}$ & 0   & 0.1 & 0.2  & 0.3  & 0.4 & 0.5 & 0.6  & 0.7  & 0.8 & 0.9  & 0.95\\
            \midrule
            mIoU& 43.7 & 43.7  & 43.7 & 43.7 & \textbf{43.8}& \textbf{43.8} & 43.7  & 43.7  & 43.4& 43.1  & 42.3 \\

            \bottomrule
        \end{tabular}
    }
    \caption{Effect of confidence threshold $Th_{p}$ for pseudo label selection in DGT-ST on Syn $\rightarrow$ Sk.}
    \label{tab:praram_thp_sac}
\end{table*}

\begin{figure*}[t]
    \centering
    \includegraphics[width=1\linewidth]{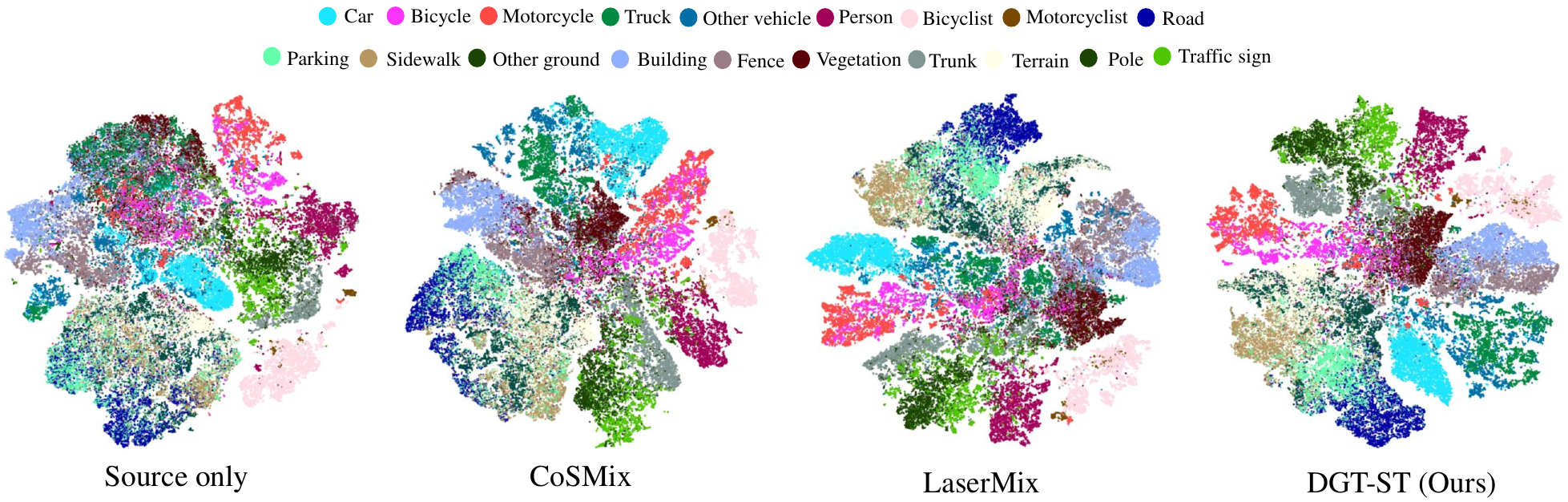}
    \caption{
        t-SNE visualization of the embedded features on Syn $\rightarrow$ Sk.
    }
    \label{fig:tsne}
\end{figure*}

\section{t-SNE visualization}
\label{sup:tsne_vis}
In \cref{fig:tsne}, we visualize the learned features of source only, CoSMix, LaserMix, and our DGT-ST by t-SNE~\cite{van2008visualizing}. We can observe that semantically similar categories are mixed together for all methods, \eg, the features of road, sidewalk, and parking are mixed, and features of pole and traffic sign are mixed.  In comparison, we can more easily separate different classes features of DGT-ST, \eg, the trunk and other classes, the pole and traffic sign classes, and the building and fence classes. Therefore, we can conclude that DGT-ST extracts more discriminative features than the other works.

\begin{figure*}
    \centering
    \includegraphics[width=1\linewidth]{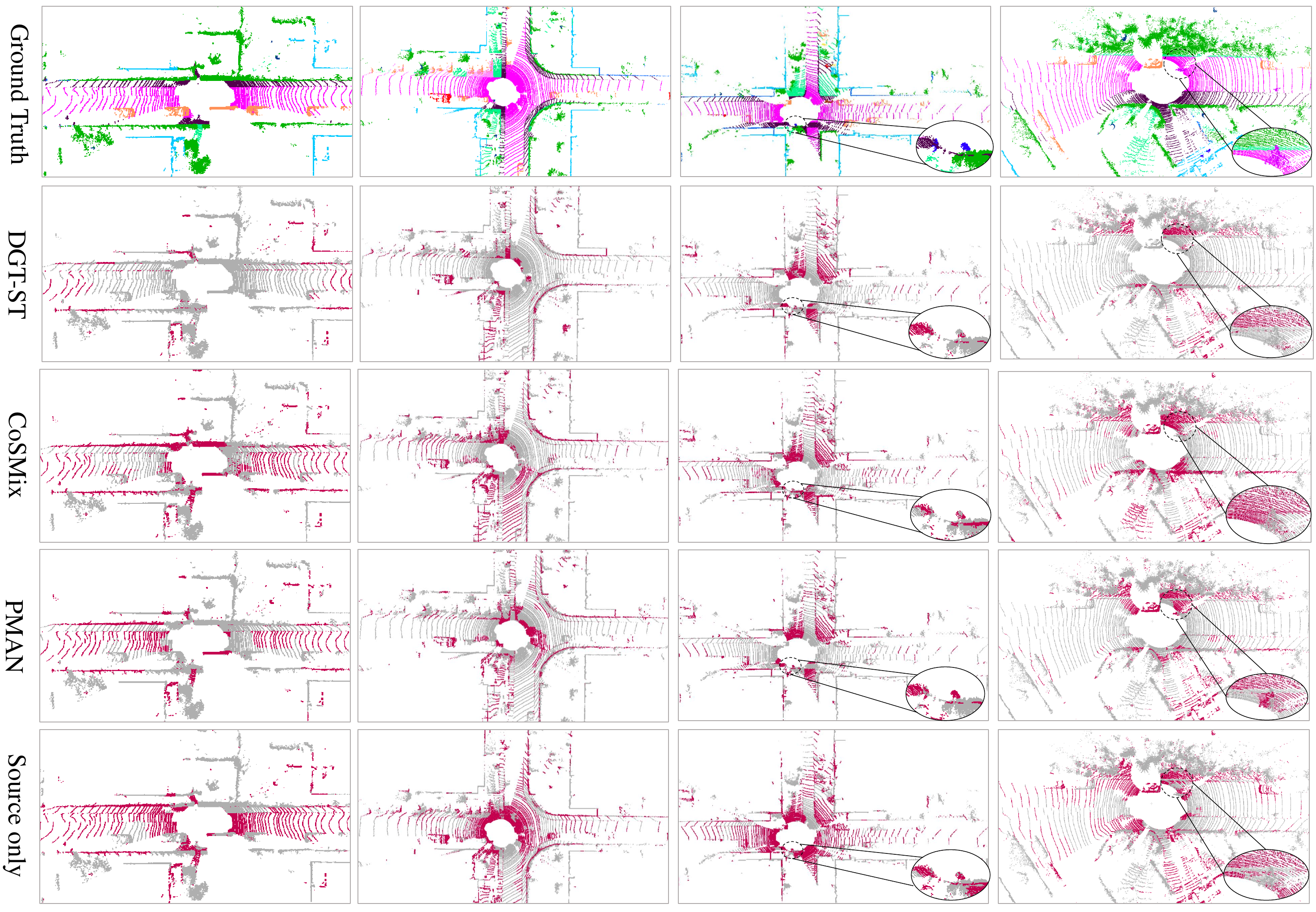}
    \caption{
        Additional qualitative results (error maps) on Syn $\rightarrow$ Sk. To highlight the differences, the correct and incorrect predictions are painted in gray and red, respectively.
    }
    \label{fig:gray_read_com}
\end{figure*}
\section{More qualitative results}
\label{sup:more_vis_com}
In \cref{fig:gray_read_com}, we present more visualization results (error maps) on Syn $\rightarrow$ Sk, and compare
our results with source only, PMAN, CoSMix, and the ground truth. Obviously, the incorrect predictions of DGT-ST are significantly fewer than other methods.

\end{document}